\documentclass[10pt,twocolumn,letterpaper]{article}

\usepackage{iccv}
\usepackage{times}
\usepackage{epsfig}
\usepackage{graphicx}
\usepackage{amsmath}
\usepackage{amssymb}
\usepackage{multirow}
\usepackage[caption=false]{subfig}
\usepackage{booktabs}
\usepackage[title]{appendix}
\usepackage[pagebackref=true,breaklinks=true,letterpaper=true,colorlinks,bookmarks=false]{hyperref}

\iccvfinalcopy 

\def\iccvPaperID{3867} 

\ificcvfinal\fi

\begin{document}

\title{See More and Know More: Zero-shot Point Cloud Segmentation via \\ Multi-modal Visual Data}

\author{\textbf{
Yuhang Lu$^{1,}$\thanks{Equal contribution. $\dagger$ Corresponding author.}  , 
Qi Jiang$^{1,}$\footnote[1]{} , 
Runnan Chen$^{2}$, 
Yuenan Hou$^{3}$, 
Xinge Zhu$^{4}$, 
Yuexin Ma$^{1,}$\footnote[2]{}}\\ 
$^{1}$ ShanghaiTech University
$^{2}$ The University of Hong Kong \\
$^{3}$ Shanghai AI Laboratory 
$^{4}$ The Chinese University of Hong Kong\\
{\tt\small \{luyh2,jiangqi,mayuexin\}@shanghaitech.edu.cn}}


\maketitle
\ificcvfinal\thispagestyle{empty}\fi

\begin{abstract}

Zero-shot point cloud segmentation aims to make deep models capable of recognizing novel objects in point cloud that are unseen in the training phase. Recent trends favor the pipeline which transfers knowledge from seen classes with labels to unseen classes without labels. They typically align visual features with semantic features obtained from word embedding by the supervision of seen classes' annotations. However, point cloud contains limited information to fully match with semantic features. In fact, the rich appearance information of images is a natural complement to the textureless point cloud, which is not well explored in previous literature. Motivated by this, we propose a novel multi-modal zero-shot learning method to better utilize the complementary information of point clouds and images for more accurate visual-semantic alignment. Extensive experiments are performed in two popular benchmarks, \ie, SemanticKITTI and nuScenes, and our method outperforms current SOTA methods with $52\%$ and $49\%$ improvement on average for unseen class mIoU, respectively.
\end{abstract}

\section{Introduction}
\label{sec:intro}
Point cloud segmentation is a critical task for 3D scene understanding, which promotes the development of autonomous driving, assistive robots, digital urban, AR/VR, etc. Fully supervised methods~\cite{zhu2020cylindrical,Xu2021RPVNetAD, Hong2021LiDARbasedPS, kong2023rethinking} have achieved impressive performance. However, there exist tremendous categories of objects in the real world, especially in large-scale outdoor scenes, bringing challenges for such methods to generalize to novel objects without labels in training data. Furthermore, manual annotations for 3D point clouds are extremely time-consuming and expensive. Zero-shot learning can recognize unseen objects by utilizing side information, especially the word embedding, to transfer the knowledge of seen categories to unseen ones, which is important for the point cloud segmentation in large-scale scenes.

\begin{figure}[t]
  \includegraphics[width=1\linewidth]{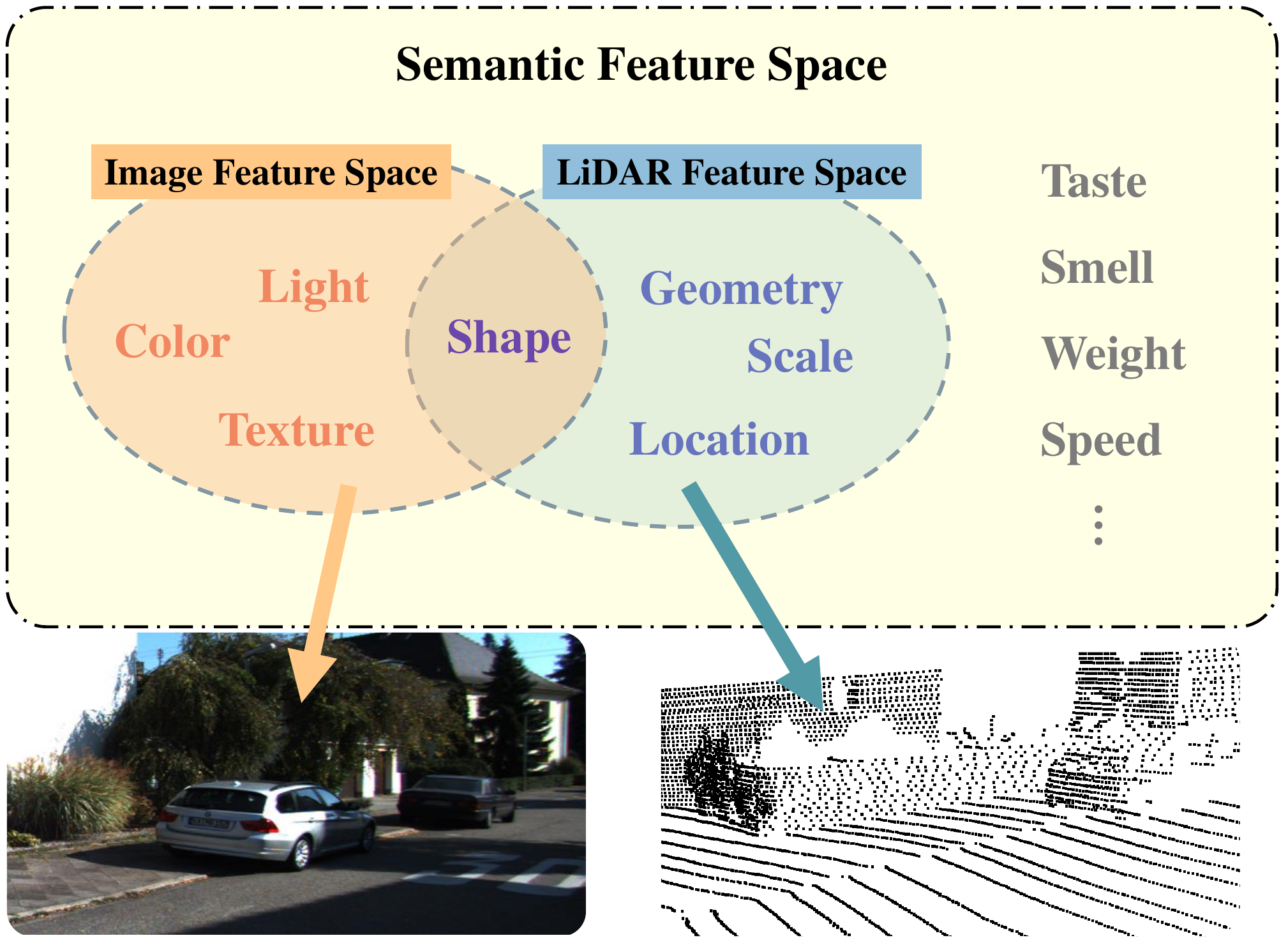}
  \caption{
  Semantic features of objects obtained by word embedding contain rich and diverse information, including appearance characteristics existing in images(\ie, color, light), geometry and location information contained in LiDAR point clouds(\ie, scale, shape), and some other non-visual properties(\ie, smell, weight). Previous image-based or point cloud-based zero-shot learning only considers the alignment between uni-modal visual features and semantic features, where the former can just match a small subset of the latter. We propose a more effective solution for zero-shot 3D segmentation by using multi-modal visual features.
  }
  \label{fig:teaser}
\end{figure}

Zero-shot semantic segmentation on 2D images has made promising progress in the past few years~\cite{Xian_2019_CVPR,joem,NEURIPS2019_0266e33d, CSRL, Cheng_2021_ICCV, Gu_2020}. There are two main streams of methods, including generative methods and projection-based methods, which inspire the following research works on 3D point clouds. For generative methods~\cite{9665941,liu2021language}, they usually train a fake feature generator supervised by seen classes and fine-tune a classifier for recognizing real seen-class features and synthesized unseen-class features. However, 3D features are more difficult to generate than 2D features due to higher dimensional information, making such strategies perform unsatisfactorily on 3D point clouds. Moreover, these methods require additional training efforts when new unseen categories appear, which limits the generalization capability on real-world applications. For projection-based approaches~\cite{chen2022zeroshot}, they target to align visual features to corresponding semantic features by the seen-class supervision, so that unseen class can be recognized by leveraging the similarity between its visual features and semantic features. Such methods can be easily generalized to novel classes without retraining. However, visual features extracted from the point cloud can only match a subset of semantic features and yield limited performance, as shown in Fig.~\ref{fig:teaser}.

In fact, current autonomous vehicles and robots are usually equipped with multiple sensors, where LiDAR and camera are the most common ones~\cite{nuscenes2019,behley2019iccv}. Since point cloud contain accurate location and geometric information and images provide rich color and texture characteristics, many researchers focus on exploring sensor-fusion methods~\cite{chen2022futr3d,liang2022bevfusion,wang2021pointaugmenting} for achieving more precise perception. Considering that see more and know more, we aim to make these two uni-modal visual data complement each other and generate more comprehensive visual features to better align with semantic features for more effective zero-shot learning. To our knowledge, we are the \textbf{first} to explore zero-shot learning based on multi-modal visual data.

In this paper, we focus on transductive generalized zero-shot learning for point cloud-based semantic segmentation, where both seen and unseen classes will appear in one scene but only objects of seen classes have labels during training. Based on the input of the synchronized point cloud and image, we propose a novel zero-shot point cloud segmentation method. Specifically, we propose an effective multi-modal feature fusion approach, termed \textbf{Semantic-Guided Visual Feature Fusion (SGVF)}, to obtain a more comprehensive visual feature representation, where valuable information from two uni-modal visual features are adaptively selected under the guidance of semantic features. As opposed to previous sensor-fusion methods, our strategy is more flexible and applicable for zero-shot learning by introducing semantic features to play an active role in the visual feature fusion stage. In this condition, exactly valid information can be utilized for the following semantic-visual feature alignment. Then, the knowledge of seen classes can be effectively transferred to unseen classes. Furthermore, to reduce the semantic-visual domain gap in advance, we propose \textbf{Semantic-Visual Feature Enhancement (SVFE)} to enhance both semantic features and visual features by transferring the domain knowledge, such as relationships among classes, to each other, which definitely benefits the following SGVF and the final semantic-visual alignment process. Actually, our method can be easily extended to more visual modalities.

We conduct extensive comparisons with current 2D and 3D zero-shot segmentation methods and our method outperforms others significantly on different datasets and settings. The effectiveness of each module of our method is also verified by ablation studies. In summary, our contributions are summarized as follows:

\begin{itemize}
\item[$\bullet$] We propose a novel multi-modal zero-shot approach for point cloud semantic segmentation.
\item[$\bullet$] We design an effective feature-fusion method with semantic-visual feature enhancement, which can better align visual features with semantic features to benefit the recognition of unseen classes.
\item[$\bullet$] Our method achieves state-of-the-art performance on SemanticKITTI and nuScenes datasets.
\end{itemize}

\section{Related Work}
\label{sec:related}
\subsection{Zero-Shot Learning}
Zero-shot learning aims at transferring the knowledge learned from seen categories to unseen ones. Many zero-shot learning studies\cite{Changpinyo2016SynthesizedCFzsl1, Kodirov2017SemanticAFzsl2, Zablocki2019ContextAwareZLzsl3, Mishra2018AGMzsl4, Lampert2009LearningTDzsl5, Akata2013LabelEmbeddingFAzsl6, Xian2016LatentEF_zsl7, Lampert2014AttributeBasedCF_zsl8, Bucher2017GeneratingVR_zsl9, Demirel2017Attributes2ClassnameAD_zsl11,Li2018DeepSS_zsl12,Gan2015ExploringSI_zsl13} leverage intermediate representations such as semantic embeddings and attributes to bridge the gap between seen and unseen classes. Early works of zero-shot learning (ZSL)~\cite{Akata2013LabelEmbeddingFAzsl6} only recognize unseen classes of data during inference. While the recently discussed generalized zero-shot learning (GZSL)~\cite{Scheirer2013TowardOS_gzsl} requires model to recognize both seen and unseen classes, which is more challenging yet practical since real scenes usually contain both seen and unseen classes of objects at the same time. Apart from ZSL and GZSL, zero-shot learning tasks can also be classified as inductive\cite{Akata2015EvaluationOO_izsl1, Changpinyo2016SynthesizedCF_izsl2, Frome2013DeViSEAD_izsl3, Norouzi2014ZeroShotLB_izsl4, Zhang2015ZeroShotLV_izsl5, RomeraParedes2015AnES_izsl6} and transductive\cite{Song2018TransductiveUE_tzsl1,Zhao2018DomainInvariantPL_tzsl2,Yu2018TransductiveZL_tzsl3, Guo2016TransductiveZR_tzsl4, Kodirov2015UnsupervisedDA_tzsl5}. The former excludes the occurrence of samples of unseen classes yet the latter permits. In this paper, we focus on the setting of transductive GZSL, which is more practical for real-world applications.

\subsection{Zero-shot Segmentation on 2D Image}\label{sec:zs3_2d}
Zero-shot segmentation on 2D images has been widely explored~\cite{Xian_2019_CVPR,joem,NEURIPS2019_0266e33d, CSRL, Cheng_2021_ICCV, Gu_2020, Kato2019ZeroShotSS_zs31, Hu2020UncertaintyAwareLF_zs32, Pastore2021ACL_zs33, Lv2020LearningUZ_zs34,Zheng2021ZeroShotIS,Xu2022GroupViTSS} in the past few years, which can be divided into projection-based methods and generative methods. Projection-based methods like SPNet\cite{Xian_2019_CVPR} intend to align visual feature space with semantic feature space so as to generalize the model to unseen data by leveraging the structure of the semantic feature space. Since the training process only involves labels of seen classes, many methods~\cite{Xian_2019_CVPR,Chao2016AnES} try to alleviate the bias toward seen classes during training.
By contrast, generative methods\cite{NEURIPS2019_0266e33d, CSRL, Cheng_2021_ICCV, Gu_2020} usually adopt a multi-stage training paradigm with a fake feature generator supervised with seen data and a classification layer fine-tuned by real seen-class features and synthesized unseen-class features. 
Following the astonishing zero-shot transfer learning performance of CLIP~\cite{CLIP}, a series of works\cite{Xu2021clip1, Lddecke2022clip2, Ding2022Decouplingclip3,Li2022LanguagedrivenSS, zhou2022extract} begin to exploit its huge potential under ZS3 task and have made significant improvement.
However, data leakage is a concern since unseen objects may already occur in the CLIP training data and it is also difficult to extend to 3D tasks due to the lack of huge 3D pre-training samples.

\subsection{Zero-shot Segmentation on Point Cloud}

The boom of autonomous driving and the expensive 3D manual annotation has led to zero-shot point cloud perception becoming an emerging research hotspot.
Many works\cite{cheraghian2019zero, cheraghian2019mitigating, cheraghian2022zero, cheraghian2020transductive, liu2021language, 9665941, chen2023labelfree, chen2023clip2scene, chen2022zeroshot} focusing on zero-shot learning for 3D point clouds appear, especially for point cloud classification~\cite{cheraghian2019zero, cheraghian2019mitigating, cheraghian2022zero, cheraghian2020transductive}. Cheraghian et al.\cite{cheraghian2019zero} uses PointNet\cite{Qi_2017_CVPR} to extract point cloud features and leverages W2V\cite{mikolov2013distributed} or Glove\cite{pennington2014glove} as extra semantic features. Then, many researchers try to address the hubness problem~\cite{cheraghian2019mitigating,cheraghian2022zero} and extend zero-shot learning to the transductive setting~\cite{cheraghian2020transductive}.  

To the best of our knowledge, only three papers\cite{liu2021language, 9665941, chen2022zeroshot} propose solutions for zero-shot point cloud semantic segmentation. Among them, 3DGenZ\cite{9665941} and SeCondPoint\cite{liu2021language} are generative approaches. They all generate fake features of unseen classes with semantic features for training the classifier to achieve the zero-shot transfer. However, generative methods require extra training efforts when new unseen categories appear, which limits the generalization capability. In addition, since 3D features are more complex than 2D features, the generated feature distribution does not fit the original distribution well, resulting in poor results. Different from them, TGP\cite{chen2022zeroshot} is a 
projection-based approach, which learns geometric primitives to facilitate the knowledge transfer from seen classes to unseen categories. 
However, only relying on the point cloud properties, the alignment between visual space and semantic space is difficult since there is a huge domain gap between these two spaces and the point cloud could only match a subset of semantic space, causing incomplete knowledge for unseen objects.

\subsection{Multi-Sensor Fusion}
Considering that the image contains rich appearance features and 
the point cloud possesses accurate location and geometry features, many works~\cite{Xu_2018_CVPR, Vora_2020_CVPR, wang2021pointaugmenting, XuYan20222DPASS2P, Zhuang_2021_ICCV, Bai_2022_CVPR, liang2022bevfusion, chen2022futr3d, li2022deepfusion} explore effective fusion ways to make these two sensors complement each other for more precise 3D perception.
PointPainting\cite{Vora_2020_CVPR} and PointAugmenting\cite{wang2021pointaugmenting} utilize the semantic label or feature at the projected image location as additional information to append to the corresponding point, while such point-level fusion strategy will lose dense appearance feature of images. PMF\cite{Zhuang_2021_ICCV} performs perspective projection on the point cloud and performs feature fusion in the camera coordinate system. 2DPASS\cite{XuYan20222DPASS2P} leverages knowledge distillation for cross-modal knowledge transfer. Recently, transformer-based sensor-fusion methods~\cite{Bai_2022_CVPR,chen2022futr3d, 9863660, li2022deepfusion} achieve promising performance for 3D perception with learnable projection and the usage of the global context. However, these feature-fusion methods are designed for fully supervised tasks. For zero-shot segmentation, direct feature fusion operation results in a more complex fused visual feature, making the alignment to semantic features more difficult. We allow semantic features to adaptively select desired features from two visual modalities for matching, avoiding the interference of irrelevant information.

\section{Methods}
\label{sec:methods}
\begin{figure*}
  \includegraphics[width=1\linewidth]{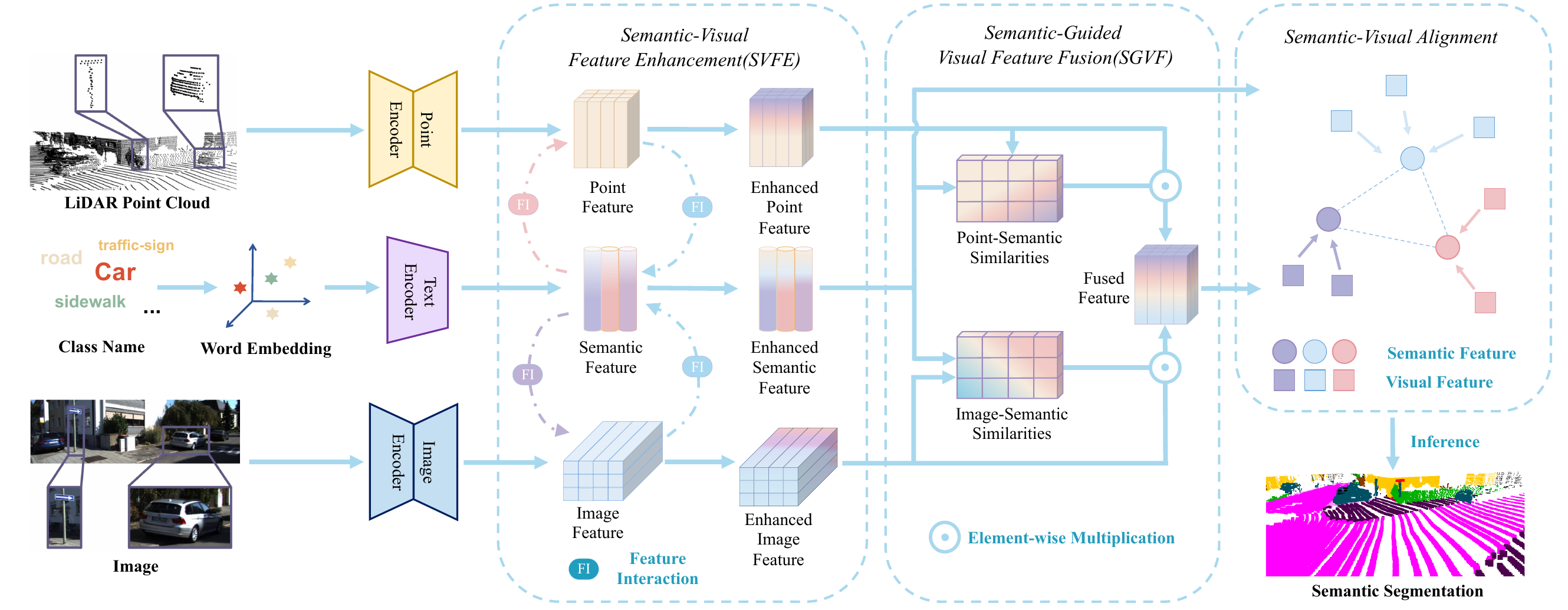}
  \caption{Method overview. Firstly, 3D and 2D backbones extract visual features from LiDAR point cloud and image, while MLP extracts semantic features. Secondly, for reducing the semantic-visual gap, visual features and semantic features interact with each other by learnable projection in the SVFE module. Then, we make semantic features adaptively select valuable visual features from two modalities for effective feature fusion in the SGVF module. Finally, we perform semantic-visual feature alignment for zero-shot learning.}
  \label{fig:main}
\end{figure*}

\subsection{Problem Formulation}
Point cloud semantic segmentation aims at classifying each point into a specified class. Similar to ~\cite{chen2022zeroshot,9665941}, we divide all classes into seen and unseen ones. We focus on the generalized transductive zero-shot point cloud segmentation problem, which is a more realistic setting where the model needs to segment both the seen and unseen classes in the scene by seeing their features and supervised by the labels of only seen classes.

Let $P \in \mathbb{R}^{T\times 3}$ denote one frame of point cloud with $T$ points represented by $(x,y,z)$ coordinates, and $X \in \mathbb{R}^{3\times \mathcal{H}\times \mathcal{W}}$ denote the corresponding image, where $\mathcal{H}\times \mathcal{W}$ means the image size. The set of seen and unseen classes are expressed as $C^{s}=\left\{c_{i}^{s}\right\}_{i=1}^{N^s}$ and $C^{u}=\left\{c_{i}^{u}\right\}_{i=1}^{N^u}$($C^{s}\cap C^{u}=\emptyset$), respectively, where $s, u$ stand for seen and unseen categories and $N^s$ and $N^u$ denote the number of data samples involving seen and unseen categories. $W^{s}=\left\{w_{i}^{s}\right\}_{i=1}^{N^s}$ and $W^{u}=\left\{w_{j}^{u}\right\}_{j=1}^{N^u}$ are the word embedding~\cite{chen2022zeroshot} of seen and unseen class names from the word2vec~\cite{Mikolov2013EfficientEO} or glove~\cite{pennington2014glove}, respectively. Since we focus on the transductive zero-shot segmentation setting, the training set is defined as $D_{train}$ = $\left \{(P_{i}^{s}, X_{i}^{s}, W_{i}^{s}, Y_{i})_{i=1}^{N^{s}}, (P_{j}^{u}, X_{j}^{u}, W_{j}^{u})_{j=1}^{N^{u}}\right \}$, where $Y$ is the ground truth label for seen categories.


\subsection{Overview}

As illustrated in Fig.~\ref{fig:main}, our method consists of four main modules, including Feature Extraction, Semantic-Visual Feature Enhancement (SVFE), Semantic-Guided Visual Feature Fusion (SGVF), and Semantic-Visual Alignment. Firstly, following~\cite{chen2022zeroshot}, we use a 3D backbone network to produce the point visual representation $F_{l}$, and utilize the 2D backbone network to extract image visual representation $F_{i}$. Meanwhile, a Multi-Layer Perception (MLP) $G(\cdot)$ is used to project the word embedding $W$ into $F_{s}$ as the semantic feature of specific categories. Secondly, to reduce huge domain gaps between visual and semantic features, we design SVFE to make these two feature space interact knowledge with each other to enhance the feature representation by the cross-attention mechanism. Then, we propose SGVF to make semantic features automatically select valuable information from two visual modalities for better feature alignment. In the end, we perform semantic-visual feature alignment for transferring the knowledge from seen objects to unseen ones for zero-shot point cloud segmentation.

In the following sections, we will introduce the technical details of modules according to the following order: SGVF, SVFE, and Semantic-Visual Alignment.


\subsection{Semantic Guided Visual Feature Fusion}

Point cloud contains precise location and geometry information, while images provide rich texture and color information. The combination of both visual features can better match semantic features extracted from language descriptions, which may contain the information of diverse properties of the category. Therefore, we propose to leverage multi-modal visual data for semantic-visual alignment to solve zero-shot learning problems. However, direct fusing point cloud feature and image feature in the data level~\cite{Vora_2020_CVPR} or feature level~\cite{wang2021pointaugmenting} will make the fused feature more complex, resulting in difficulty in aligning with semantic features. We design an adaptive selection mechanism for semantic features, where the network can learn valuable information from two visual modalities automatically under the semantic guidance and integrate them together as the richer visual feature.

Based on the point cloud feature $F_{el}$, image feature $F_{ei}$, and semantic feature $F_{es}$ gained from the last module (Section.~\ref{sec.svfe}), we search valuable visual features for semantic features from the 3D point cloud and 2D image by calculating the weight matrix $w_{3D}$ and $w_{2D}$, respectively. It is conducted by utilizing the multi-head attention~\cite{Transformer}: 
\begin{equation}
\begin{aligned}
&w_{3D} = \mathrm{MultiHeadAttention}(F_{es}, F_{el}),\\
&w_{2D} = \mathrm{MultiHeadAttention}(F_{es}, F_{ei}).
\end{aligned}
\label{func:weight}
\end{equation}
The weights stand for the significance of two uni-modal visual features to semantic features. And the fused visual feature can be obtained by applying element-wise multiplication between the weight matrix and visual features. Then, we obtain the final fused visual feature by employing an MLP:
\begin{equation}
\begin{aligned}
&F_{fusion} = \mathrm{MLP}(\mathrm{softmax}(\mathrm{stack}(w_{3D}\odot F_{el}, w_{2D}\odot F_{ei}))).
\end{aligned}
\label{func:fusion}
\end{equation}

In this way, the network utilizes multi-modal visual data effectively by selecting valuable information to match semantic features for different categories of objects, which can benefit the following alignment of semantic and visual spaces, thus improving the recognition ability of unseen objects.



\subsection{Semantic-Visual Feature Enhancement}
\label{sec.svfe}

During the selection step in SGVF, the huge domain gap between visual features and semantic features will hinder the learning process for fusing effective visual features. Therefore, we consider narrowing the semantic-visual gap in advance by transferring the knowledge, such as relationships among various categories, between semantic and visual space. We conduct the knowledge interaction by the cross-attention mechanism, which can learn the semantic-visual projection automatically and enhance each feature with valuable knowledge of the other.


\textbf{Semantic Feature Enhancement.} To enhance the semantic feature $F_s$ by visual features, we take $F_s$ as the query $q$ and visual feature as key $k$ and value $v$ to feed into a Transformer Decoder~\cite{Transformer} as follows.
\begin{equation}
\begin{aligned}
&\mathrm{TD}(q,k,v) = \mathrm{Linear}(\mathrm{LN}(\mathrm{MLP}(Q) + Q)),\\
&Q = \mathrm{LN}(\mathrm{CrossAttention}(q, k, v) + q),
\end{aligned}
\end{equation}
where $\mathrm{Linear}$ indicates linear mapping layer and $\mathrm{LN}$ denotes layer normalization. Because we have two modalities of visual features, we make them interact with the semantic features in order. Considering that we target for point cloud segmentation, we first enhance feature $F_s$ by point feature $F_l$ to pull the representation of two spaces closer, and then conduct the same operation on image feature $F_i$ for further semantic feature enhancement. Formula.~\ref{func:transdecoder} illustrates the process. 
\begin{equation}
\begin{aligned}
&F_{es} = \mathrm{TD}(\mathrm{TD}(F_{s}, F_{l},F_{l}),F_{i},F_{i}).
\end{aligned}
\label{func:transdecoder}
\end{equation}

\textbf{Visual Feature Enhancement.} Similarly, we enhance the visual feature by querying to semantic feature and fetching knowledge from semantic space. Then, we obtain enhanced point feature by $F_{el} = \mathrm{TD}(F_{l}, F_{s}, F_{s})$, and enhanced image feature by $F_{ei} = \mathrm{TD}(F_{i}, F_{s}, F_{s})$.


In this way, we reduce the difference between semantic and visual rpresentations by feature interaction and the enhanced features can further facilitate the visual feature selection process in SGVF and the final alignment between semantic and visual spaces. 

\begin{figure}[t]
\centering
	\subfloat[Before SVFE]{\includegraphics[width = 0.155\textwidth]{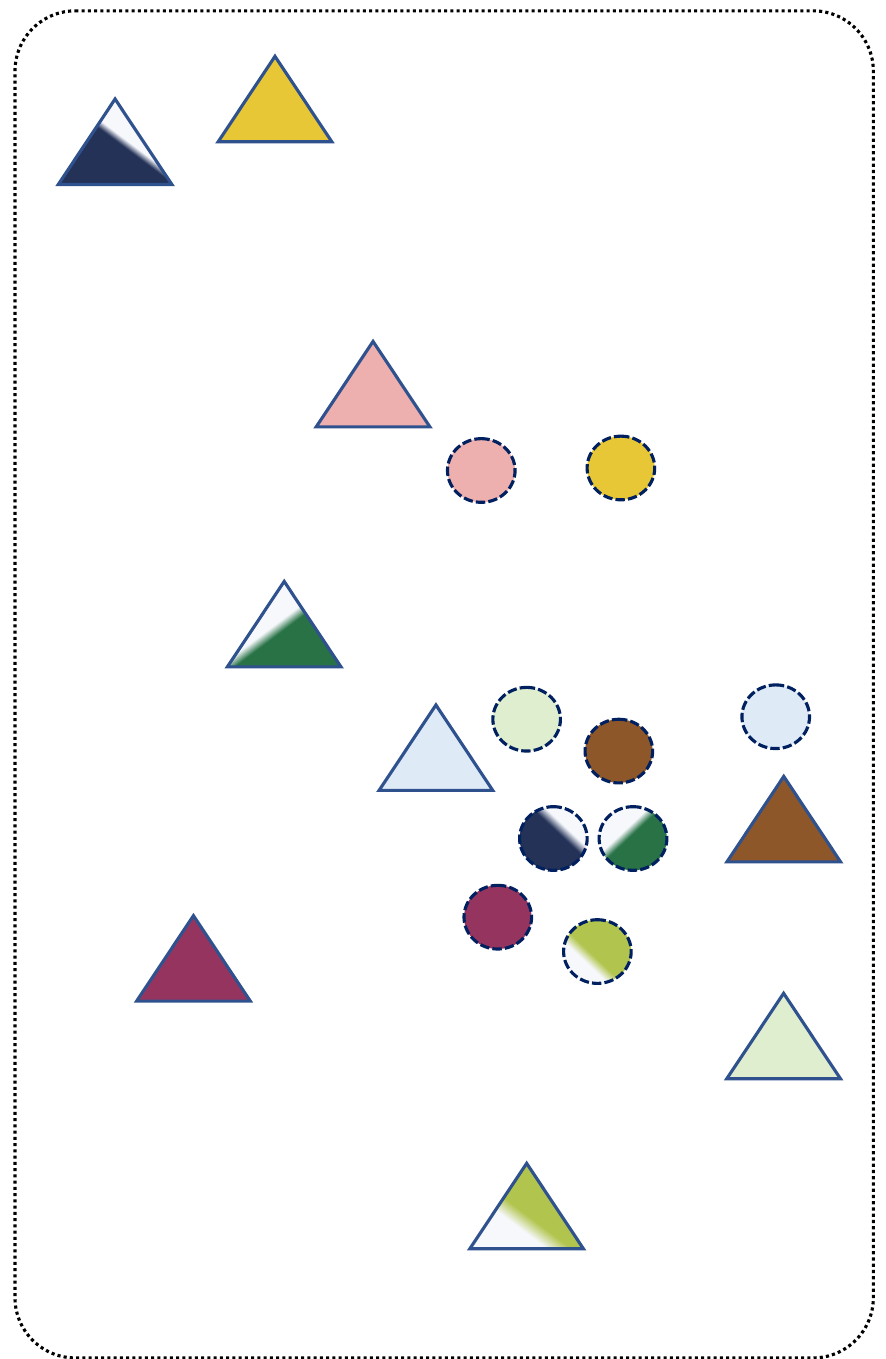}}
	\hfill
	\subfloat[After SVFE]{\includegraphics[width = 0.155\textwidth]{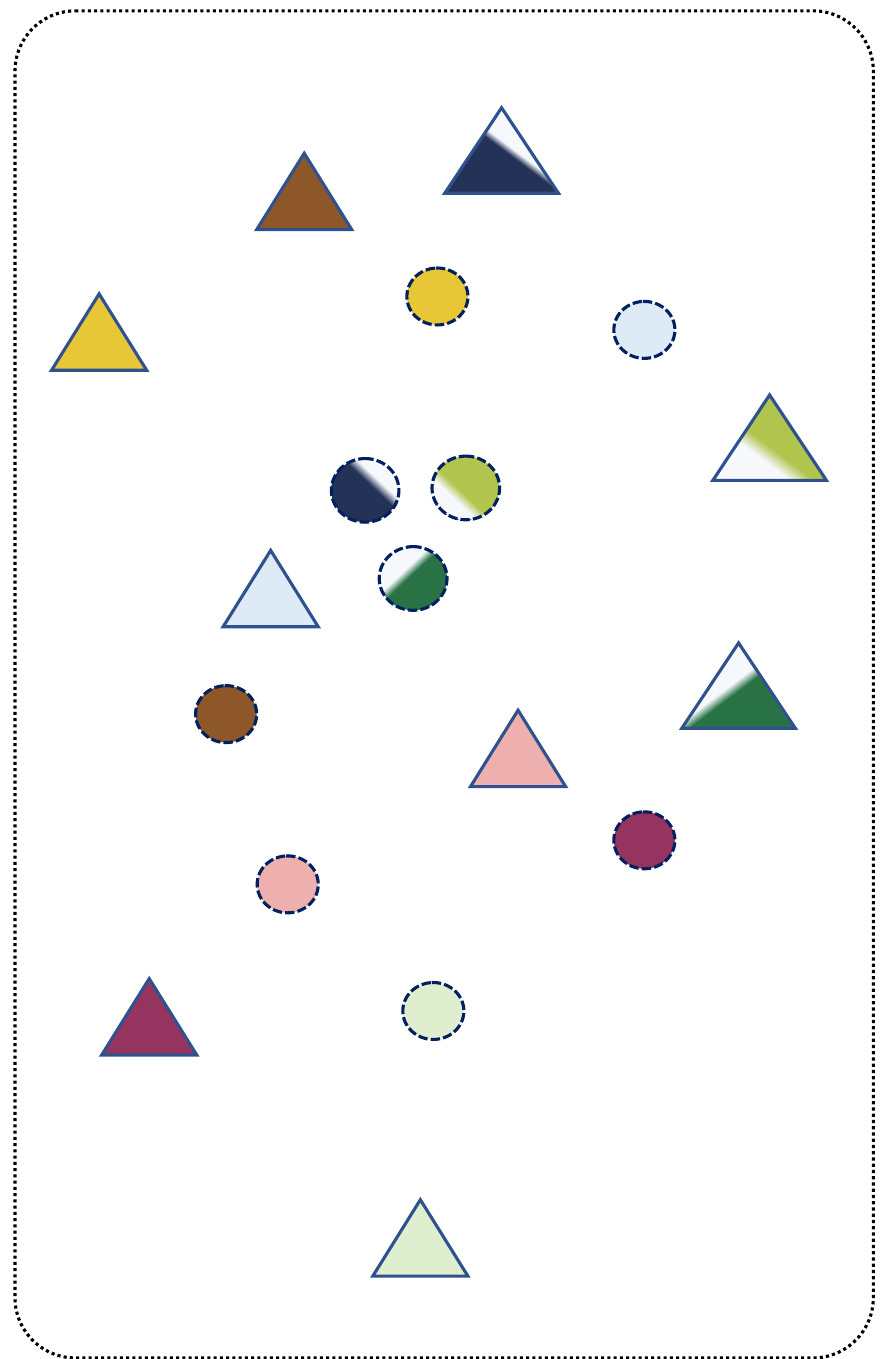}}
	\hfill
	\subfloat[After SGVF]{\includegraphics[width = 0.155\textwidth]{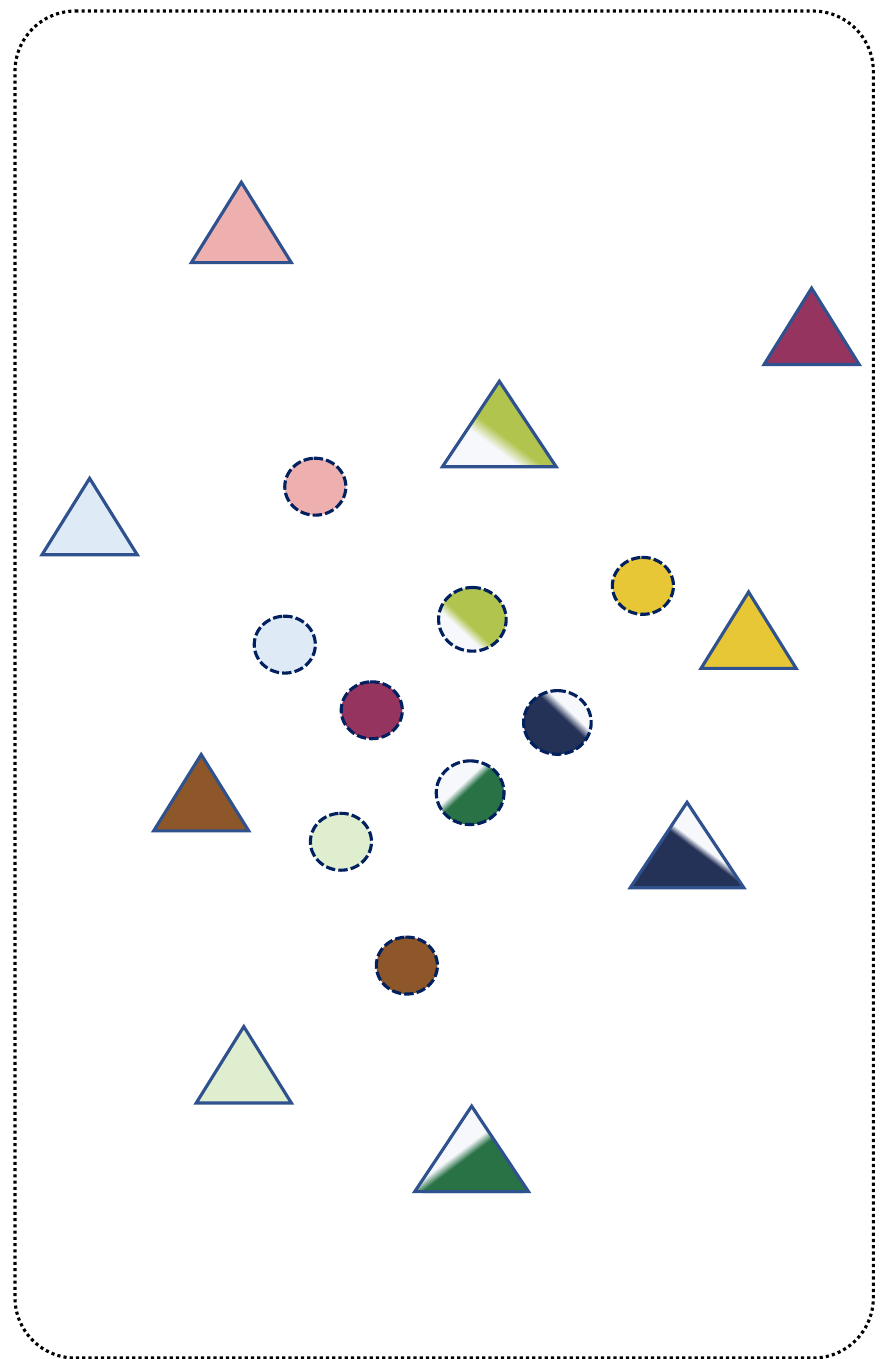}}
	\hfill
\caption{
Visualization of semantic-visual feature relationships of various classes in one scene of SemanticKITTI by t-SNE. Triangles indicate semantic features, circles denote visual features, pure colors stand for seen classes, and gradient colors mean unseen classes. The same color represents the same class. Visual features in (a) and (b) come from the point cloud branch. 
}
\label{Fig:vs-relationship-vis}
\end{figure}

To demonstrate the effectiveness of SVFE and SGVF intuitively, we select one scene from SemanticKITTI validation set and visualize semantic and visual features of all classes occurred in Fig.~\ref{Fig:vs-relationship-vis}. It is obvious to see that our model pulls visual features to corresponding semantic features gradually by effective semantic-visual feature enhancement and multi-modal visual feature fusion.

\subsection{Semantic-Visual Alignment}



Through the feature enhancement of SVFE and multi-modal visual feature fusion of SGVF, we obtain a comprehensive representation of visual features, which can match more content and represent similarly with semantic features. We align visual and semantic feature spaces by the supervision of seen classes. Therefore, the knowledge of seen class can be transferred to unseen class with the aid of side information, e.g., semantic features from word embedding.

\textbf{Loss function. }Following TGP\cite{chen2022zeroshot}, we adopt a cross entropy loss and an unknown-aware InfoNCE loss to distinguish various seen classes and allow the model to identify whether an object is a seen class or an unseen class. 

In order to obtain a better semantic-visual space alignment, we have to ensure that the distribution of each class is compact within classes and distinguishable between classes. To this end, we use the following objective function.
\begin{equation}
\begin{aligned}
L_{s} = -\log \sum_{i}^{N^{s}} \sum_{t}^{T_{i}}\frac{\mathrm{exp}(D(f_{i}^{t}, e_{y_{i}^{t}})/\tau )}
{ {\textstyle \sum_{c=1}^{C^s+C^u}} \mathrm{exp}(D(f_{i}^{t}, e_{c})/\tau)},  
\end{aligned}
\end{equation}
where $f_{i}^{t}$ denotes the visual features of the $t$-th point in the $i$-th sample, $e_{y_{i}^{t}}$ is the corresponding ground truth semantic representation. $\tau$ is the inversed temperature term. $C^s$ and $C^u$ are the number of seen and unseen classes, respectively. $D$($\cdot$) is the similarity function between visual and semantic features. In this paper, we choose the dot product similarity.

Since only seen classes have annotations during training, the zero-shot model is naturally biased towards the seen classes. To avoid this, we push the features of the seen and unseen classes apart by the following loss formula.
\begin{equation}
\begin{aligned}
L_{u} = \log \sum_{j}^{N^{u}} \sum_{t}^{T_{j}}\frac{ {\textstyle \sum_{c=1}^{C^s}\mathrm{exp}(D(f_{j}^{t}, e_{c})/\tau}) }
{ {\textstyle \sum_{\hat{c}=1}^{C^s+C^u}} \mathrm{exp}(D(f_{j}^{t}, e_{\hat{c}})/\tau)}.
\end{aligned}
\end{equation}
The overall loss function is the combination of two losses.
\begin{equation}
\begin{aligned}
L = L_{s} + L_{u}.
\end{aligned}
\end{equation}

\textbf{Inference. }We infer a new scene with fused visual feature $F_{fusion}$ = $\left \{ f_{fusion}^{t} \right \} _{t=1}^{T}$ and semantic feature $F_{es} = \left \{ e_{c} \right \}_{c=1}^{C^s+C^u}$, where $T$ is the number of points in this scene. The class of the $t$th point is determined as follows.
\begin{equation}
\begin{aligned}
C_{t} = \mathop{\arg\max}\limits_{c}\frac{\exp{(D(f_{fusion}^{t},e_{c}))}}{ {\textstyle \sum_{\hat{c}=1}^{C^s+C^u}}\exp{(D(f_{fusion}^{t},e_{\hat{c}}))} }. 
\end{aligned}
\end{equation}

\section{Experiments}
\label{sec:exp}
\begin{table*}[ht]
\caption{Comparison with state-of-the-art methods on SemanticKITTI and nuScenes datasets. We show the performance of diverse unseen-class settings introduced in Section.~\ref{sec:dataset}. Setting ``0'' indicates fully supervised manner. ``Improvement'' means the percentage improvement in the metric unseen mIoU for our method relative to the previous SOTA method. ``Supervised'' denotes that both seen and unseen classes have labels during the training of our method, which stands for the upper bound for zero-shot learning performance.} 
\resizebox{\linewidth}{!}{
\begin{tabular}{l|l|ccccc|ccccc}
\hline
\multirow{3}{*}{Setting} & \multirow{3}{*}{Model}    & \multicolumn{5}{c|}{SemanticKITTI} & \multicolumn{5}{c}{nuScenes} \\ \cline{3-12} 
 &
   &
  \multirow{2}{*}{\begin{tabular}[c]{@{}c@{}}Seen\\ mIoU\end{tabular}} &
  \multirow{2}{*}{\begin{tabular}[c]{@{}c@{}}Uneen\\ mIoU\end{tabular}} &
  \multirow{2}{*}{Improvement} &
  \multicolumn{2}{c|}{Overall} &
  \multirow{2}{*}{\begin{tabular}[c]{@{}c@{}}Seen\\ mIoU\end{tabular}} &
  \multirow{2}{*}{\begin{tabular}[c]{@{}c@{}}Uneen\\ mIoU\end{tabular}} &
  \multirow{2}{*}{Improvement} &
  \multicolumn{2}{c}{Overall} \\ \cline{6-7} \cline{11-12} 
                         &                           &      &     &     & mIoU   & hIoU   &    &    &    & mIoU  & hIoU  \\ \hline
  \multirow{2}{*}{0}       
  & TGP\cite{chen2022zeroshot}             & -    & -    & -        & 59.1 & -    & -    & -    & -       &  67.9    & -    \\
     & Ours             & -    & -    & -        & \textbf{62.6} & -    & -    & -    & -  & \textbf{69.1} & -    \\ \hline
 &
  3DGenZ\cite{9665941} &
  \multicolumn{1}{c}{40.9} &
  \multicolumn{1}{c}{12.4} &
  \multicolumn{1}{c}{-} &
  \multicolumn{1}{c}{37.9} &
  19.0 &
  \multicolumn{1}{c}{\textbf{67.8}} &
  \multicolumn{1}{c}{4.2} &
  \multicolumn{1}{c}{-} &
  \multicolumn{1}{c}{\textbf{59.9}} &
  7.9 \\ 
 &
  TGP\cite{chen2022zeroshot} &
  \multicolumn{1}{c}{58.3} &
  \multicolumn{1}{c}{28.8} &
  \multicolumn{1}{c}{+3.5\%} &
  \multicolumn{1}{c}{55.2} &
  38.6 &
  \multicolumn{1}{c}{58.9} &
  \multicolumn{1}{c}{26.9} &
  \multicolumn{1}{c}{+25.3\%} &
  \multicolumn{1}{c}{54.9} &
  36.9 \\ 
 &
  Ours &
  \multicolumn{1}{c}{\textbf{59.5}} &
  \multicolumn{1}{c}{\textbf{29.8}} &
  \multicolumn{1}{c}{-} &
  \multicolumn{1}{c}{\textbf{56.4}} &
  \textbf{39.7} &
  \multicolumn{1}{c}{59.4} &
  \multicolumn{1}{c}{\textbf{33.7}} &
  \multicolumn{1}{c}{-} &
  \multicolumn{1}{c}{56.2} &
  \textbf{43.0} \\ 
\multirow{-4}{*}{2} &
  \textcolor[RGB]{169,169,169}{Supervised} &
  \multicolumn{1}{c}{\textcolor[RGB]{169,169,169}{61.5}} &
  \multicolumn{1}{c}{\textcolor[RGB]{169,169,169}{71.8}} &
  \multicolumn{1}{c}{\textcolor[RGB]{169,169,169}{-}} &
  \multicolumn{1}{c}{\textcolor[RGB]{169,169,169}{62.6}} &
  \textcolor[RGB]{169,169,169}{66.3} &
  \multicolumn{1}{c}{\textcolor[RGB]{169,169,169}{70.1}} &
  \multicolumn{1}{c}{\textcolor[RGB]{169,169,169}{61.9}} &
  \multicolumn{1}{c}{\textcolor[RGB]{169,169,169}{-}} &
  \multicolumn{1}{c}{\textcolor[RGB]{169,169,169}{69.1}} &
  \textcolor[RGB]{169,169,169}{65.7} \\ \hline
 &
  3DGenZ\cite{9665941} &
  \multicolumn{1}{c}{41.4} &
  \multicolumn{1}{c}{10.8} &
  \multicolumn{1}{c}{-} &
  \multicolumn{1}{c}{35.0} &
  17.1 &
  \multicolumn{1}{c}{\textbf{67.2}} &
  \multicolumn{1}{c}{3.1} &
  \multicolumn{1}{c}{-} &
  \multicolumn{1}{c}{51.2} &
  5.9 \\ 
 & 
  TGP\cite{chen2022zeroshot} &
  \multicolumn{1}{c}{54.6} &
  \multicolumn{1}{c}{17.3} &
  \multicolumn{1}{c}{{ +54.9\%}} &
  \multicolumn{1}{c}{46.7} &
  26.3 &
  \multicolumn{1}{c}{65.7} &
  \multicolumn{1}{c}{14.8} &
  \multicolumn{1}{c}{{ +56.1\%}} &
  \multicolumn{1}{c}{53.0} &
  24.2 \\ 
 &
  Ours &
  \multicolumn{1}{c}{\textbf{58.8}} &
  \multicolumn{1}{c}{\textbf{26.8}} &
  \multicolumn{1}{c}{-} &
  \multicolumn{1}{c}{\textbf{52.1}} &
  \textbf{36.8} &
  \multicolumn{1}{c}{66.4} &
  \multicolumn{1}{c}{\textbf{23.1}} &
  \multicolumn{1}{c}{-} &
  \multicolumn{1}{c}{\textbf{55.6}} &
  \textbf{34.3} \\ 
\multirow{-4}{*}{4} &
  \textcolor[RGB]{169,169,169}{Supervised} &
  \multicolumn{1}{c}{\textcolor[RGB]{169,169,169}{60.3}} &
  \multicolumn{1}{c}{\textcolor[RGB]{169,169,169}{71.2}} &
  \multicolumn{1}{c}{\textcolor[RGB]{169,169,169}{-}} &
  \multicolumn{1}{c}{\textcolor[RGB]{169,169,169}{62.6}} &
  \textcolor[RGB]{169,169,169}{65.3} &
  \multicolumn{1}{c}{\textcolor[RGB]{169,169,169}{71.9}} &
  \multicolumn{1}{c}{\textcolor[RGB]{169,169,169}{60.6}} &
  \multicolumn{1}{c}{\textcolor[RGB]{169,169,169}{-}} &
  \multicolumn{1}{c}{\textcolor[RGB]{169,169,169}{69.1}} &
  \textcolor[RGB]{169,169,169}{65.8} \\ \hline
 &
  3DGenZ\cite{9665941} &
  \multicolumn{1}{c}{40.3} &
  \multicolumn{1}{c}{6.5} &
  \multicolumn{1}{c}{-} &
  \multicolumn{1}{c}{29.6} &
  11.2 &
  \multicolumn{1}{c}{53.8} &
  \multicolumn{1}{c}{3.2} &
  \multicolumn{1}{c}{-} &
  \multicolumn{1}{c}{34.8} &
  6.0 \\ 
 &
  TGP\cite{chen2022zeroshot} &
  \multicolumn{1}{c}{53.6} &
  \multicolumn{1}{c}{13.3} &
  \multicolumn{1}{c}{+79.7\%} &
  \multicolumn{1}{c}{40.9} &
  21.3 &
  \multicolumn{1}{c}{\textbf{68.8}} &
  \multicolumn{1}{c}{14.1} &
  \multicolumn{1}{c}{+56.7\%} &
  \multicolumn{1}{c}{48.3} &
  23.4 \\ 
 &
  Ours &
  \multicolumn{1}{c}{\textbf{56.6}} &
  \multicolumn{1}{c}{\textbf{23.9}} &
  \multicolumn{1}{c}{-} &
  \multicolumn{1}{c}{\textbf{46.3}} &
  \textbf{33.6} &
  \multicolumn{1}{c}{66.8} &
  \multicolumn{1}{c}{\textbf{22.1}} &
  \multicolumn{1}{c}{-} &
  \multicolumn{1}{c}{\textbf{50.0}} &
  \textbf{33.2} \\ 
\multirow{-4}{*}{6} &
  \textcolor[RGB]{169,169,169}{Supervised} &
  \multicolumn{1}{c}{\textcolor[RGB]{169,169,169}{56.8}} &
  \multicolumn{1}{c}{\textcolor[RGB]{169,169,169}{75.3}} &
  \multicolumn{1}{c}{\textcolor[RGB]{169,169,169}{-}} &
  \multicolumn{1}{c}{\textcolor[RGB]{169,169,169}{62.6}} &
  \textcolor[RGB]{169,169,169}{64.8} &
  \multicolumn{1}{c}{\textcolor[RGB]{169,169,169}{74.5}} &
  \multicolumn{1}{c}{\textcolor[RGB]{169,169,169}{60.1}} &
  \multicolumn{1}{c}{\textcolor[RGB]{169,169,169}{-}} &
  \multicolumn{1}{c}{\textcolor[RGB]{169,169,169}{69.1}} &
  \textcolor[RGB]{169,169,169}{66.5} \\ \hline
 &
  3DGenZ\cite{9665941} &
  \multicolumn{1}{c}{38.3} &
  \multicolumn{1}{c}{1.3} &
  \multicolumn{1}{c}{-} &
  \multicolumn{1}{c}{22.7} &
  2.5 &
  \multicolumn{1}{c}{36.5} &
  \multicolumn{1}{c}{2.1} &
  \multicolumn{1}{c}{-} &
  \multicolumn{1}{c}{19.3} &
  4.0 \\ 
 &
  TGP\cite{chen2022zeroshot} &
  \multicolumn{1}{c}{\textbf{53.2}} &
  \multicolumn{1}{c}{8.6} &
  \multicolumn{1}{c}{+70.9\%} &
  \multicolumn{1}{c}{\textbf{34.4}} &
  14.8 &
  \multicolumn{1}{c}{\textbf{68.4}} &
  \multicolumn{1}{c}{13.7} &
  \multicolumn{1}{c}{+56.9\%} &
  \multicolumn{1}{c}{41.1} &
  22.8 \\ 
 &
  Ours &
  \multicolumn{1}{c}{46.0} &
  \multicolumn{1}{c}{\textbf{14.7}} &
  \multicolumn{1}{c}{-} &
  \multicolumn{1}{c}{32.8} &
  \textbf{22.3} &
  \multicolumn{1}{c}{68.2} &
  \multicolumn{1}{c}{\textbf{21.5}} &
  \multicolumn{1}{c}{-} &
  \multicolumn{1}{c}{\textbf{44.9}} &
  \textbf{32.7} \\ 
\multirow{-4}{*}{8} &
  \textcolor[RGB]{169,169,169}{Supervised} &
  \multicolumn{1}{c}{\textcolor[RGB]{169,169,169}{52.1}} &
  \multicolumn{1}{c}{\textcolor[RGB]{169,169,169}{77.1}} &
  \multicolumn{1}{c}{\textcolor[RGB]{169,169,169}{-}} &
  \multicolumn{1}{c}{\textcolor[RGB]{169,169,169}{62.6}} &
  \textcolor[RGB]{169,169,169}{62.2} &
  \multicolumn{1}{c}{\textcolor[RGB]{169,169,169}{73.5}} &
  \multicolumn{1}{c}{\textcolor[RGB]{169,169,169}{64.7}} &
  \multicolumn{1}{c}{\textcolor[RGB]{169,169,169}{-}} &
  \multicolumn{1}{c}{\textcolor[RGB]{169,169,169}{69.1}} &
  \textcolor[RGB]{169,169,169}{68.8} \\ \hline
\end{tabular}}
\label{table:results}
\end{table*}

We first introduce the datasets, evaluation metrics, and implementation details. Then we show results and analysis of extensive comparison experiments and ablation studies to verify the effectiveness and superiority of our method.

\subsection{Dataset and Category Division}
\label{sec:dataset}

\textbf{SemanticKITTI} \cite{behley2019iccv} 
contains 22 sequences, where ten sequences are for training, sequence 08 for validation, and the remaining sequences are used for testing. It 
has annotations for 20 classes in total. For a full evaluation, we conduct diverse zero-shot settings with different numbers of unseen classes, including \textbf{2}-motorcycle/truck, \textbf{4}-bicyclist/traffic-sign, \textbf{6}-car/terrain, \textbf{8}-vegetation/sidewalk. The classes in the unseen set increase incrementally for different settings. Especially, the setting \textbf{4} with motorcycle, truck, bicyclist, and traffic-sign is following \cite{chen2022zeroshot} and is taken as the main setting for ablation study.

\textbf{nuScenes} \cite{nuscenes2019} 
contains 40157 annotated samples with 6 monocular camera images with $360^{\circ}$ FoV and a 32-beam LiDAR scan. It 
has annotations for 17 classes in total. We conduct several zero-shot settings with different numbers of unseen classes, including the \textbf{2}-Motorcycle/trailer, \textbf{4}-terrain/traffic-cone, \textbf{6}-bicycle/car, \textbf{8}-vegetation/sidewalk. The classes in the unseen set increase incrementally for different settings. The rest classes are taken as seen classes.

\begin{figure*}[t]
\centering
 \includegraphics[width=\textwidth]{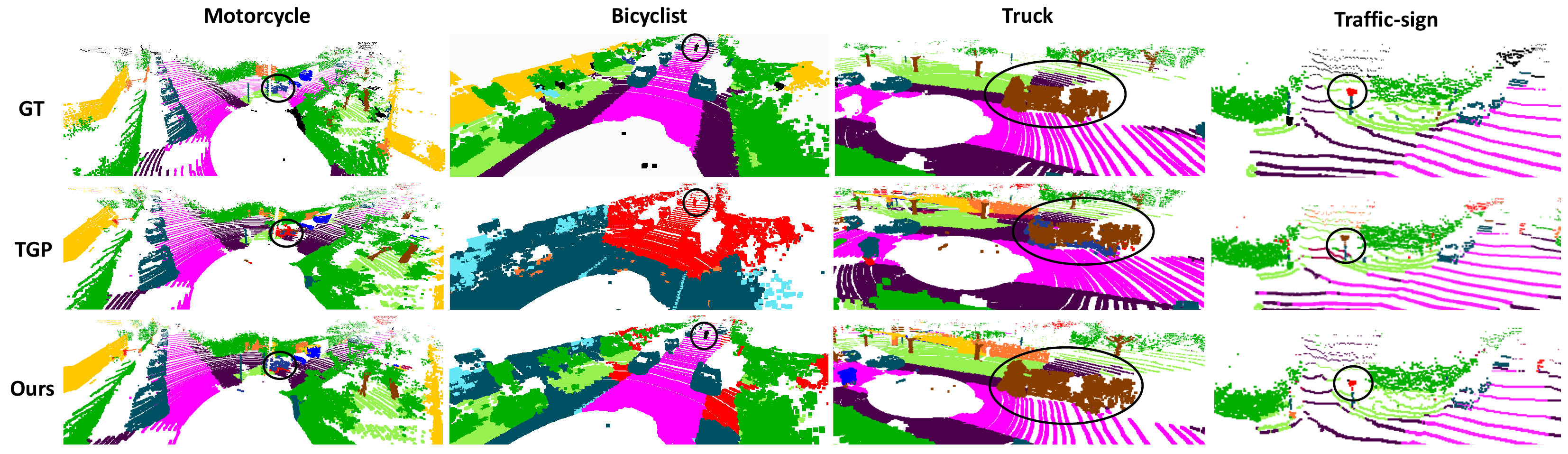}
\caption{
Visualization of results on SemanticKITTI. We show the ground truth, segmentation results of TGP\cite{chen2022zeroshot}, and segmentation results of our method in rows. Objects highlighted by black circles are unseen classes, including motorcycle, bicyclist, truck, and traffic-sign. It is obvious that our model classifies unseen classes more accurately and is closer to ground truth. 
}
\label{Fig:visualization}
\end{figure*}

\subsection{Evaluation Metrics}
We report the mean-intersection-over-union(mIoU) of seen classes, unseen classes, and all classes, respectively. Following \cite{chen2022zeroshot}, we utilize the harmonic mean IoU (hIoU) to demonstrate the overall performance of methods.
\begin{equation}
\begin{aligned}
\mathrm{hIoU} = \frac{2\times \mathrm{mIoU}_{seen}\times \mathrm{mIoU}_{unseen}}{\mathrm{mIoU}_{seen} + \mathrm{mIoU}_{unseen}},
\end{aligned}
\end{equation}
where $\mathrm{mIoU}_{seen}$ and $\mathrm{mIoU}_{unseen}$ represents the $\mathrm{mIoU}$ of seen classes and unseen classes, respectively.

\subsection{Implementation Details}
Following \cite{chen2022zeroshot}, we use Cylinder3D \cite{zhu2020cylindrical} to extract LiDAR point cloud features and ResUnet to extract image features. The visual feature dimension is 128. We adopt W2V \cite{mikolov2013distributed} and Glove \cite{pennington2014glove} to embed the class name and obtain the word embedding features (600-dimensional vector) as the auxiliary information for zero-shot segmentation. The $G(\cdot)$ is a two-layer MLP with the dimension of 96 and 128, then we obtain a 128-dimensional semantic feature. For SVFE, the transformer decoder is comprised of cross-attention and MLP. We use one decoder and the number of heads is 4. For SGVF, we utilize multi-head attention (the number of heads is 4) to compute the similarity between semantic features and visual features. Our method is built on the Pytorch platform, optimized by Adam. The learning rate for the backbone is 0.001, while the learning rate for SVFE and SGVF is 0.0003. The batch sizes for both SemanticKITTI and nuScenes are 4. It costs 80 hours to train 40 epochs on four RTX 3090 GPUs for the SemanticKITTI dataset and costs 45 hours to train 20 epochs for the nuScenes dataset.

\subsection{Comparison Results}
In this section, we show results of our method and compare it with current state-of-the-art 3D zero-shot segmentation methods, including 3DGenZ\cite{9665941} and TGP\cite{chen2022zeroshot}. Extensive experiments with different unseen class settings are conducted on SemanticKITTI and nuScenes datasets to comprehensively evaluate the methods' performance. Qualitative analysis is also provided. In particular, due to limited works on 3D zero-shot segmentation, we also adapt 2D SOTA methods to solve 3D tasks for further verification. In addition, since our method is based on multimodal fusion, we also compare with current popular multimodal fusion strategies used for full supervised tasks. Our method outperforms others in recognizing unseen classes of objects by a large margin. Detailed analysis is as follows.

\begin{table}[t]
\scriptsize
\caption{Comparison with state-of-the-art 2D methods on SemanticKITTI dataset with 4-unseen-class setting. We extend those 2D zero-shot methods to solve 3D zero-shot point cloud segmentation, where ``$\star$'' indicates results reported in \cite{chen2022zeroshot}.}
\label{table:2D_compare}
\resizebox{\linewidth}{!}{
\begin{tabular}{@{}lcccc}
\hline
\multirow{2}{*}{Model}       & \multirow{2}{*}{\begin{tabular}[c]{@{}c@{}}Seen\\ mIoU\end{tabular}} & \multirow{2}{*}{\begin{tabular}[c]{@{}c@{}}Unseen\\ mIoU\end{tabular}} & \multicolumn{2}{c}{Overall}   \\ \cline{4-5}  & & & \multicolumn{1}{c}{mIoU} & \multicolumn{1}{c}{hIoU} \\ \hline
SPNet\cite{Xian_2019_CVPR}         & 57.0 & 0.0 & 45.0 & 0.0   \\
ZS5Net$\star$ \cite{NEURIPS2019_0266e33d} & 53.2 & 5.1 & 43.1 & 9.3 \\
PMOSR$\star$ \cite{Zhang2021PrototypicalMA}       & 55.1 & 8.7 & 45.3 & 15.0 \\
JoEm \cite{Baek_2021_ICCV}         & 56.7 & 2.8 & 45.4 & 5.3 \\
\midrule
Ours                    & \textbf{58.8} & \textbf{26.8} & \textbf{52.1} & \textbf{36.8} \\ 
\bottomrule
\end{tabular}
}
\end{table}

\textbf{Comparision with 3D methods.} As shown in Table.~\ref{table:results}, our method achieves SOTA performance on 2-, 4-, 6- and 8- unseen-class settings on both SemanticKITTI and nuScenes datasets, outperforming previous SOTA methods with improvement rates of $52\%$ and $49\%$ in average for unseen class mIoU, respectively. For the comprehensive evaluation metric hIoU concerning both seen and unseen classes, our method is also superior to others by a large margin. Qualitative results are provided in Figure \ref{Fig:visualization}, where we visualize the segmentation results of our method and TGP\cite{chen2022zeroshot} on SemanticKITTI dataset alongside the ground truth annotations. Our method makes more accurate predictions on unseen classes. For example, TGP identifies a motorcycle as a traffic sign (first column) and takes parts of a truck as cyclist (third column), while our model accurately segments corresponding categories. It illustrates that multi-modal visual feature in our method really benefits the semantic-visual matching and further boost the unseen class recognition. 

\textbf{Comparision with extensions of 2D methods.}
Because 3D zero-shot segmentation just gets noticed recently, there are limited research works for comparison. Meanwhile, 2D zero-shot segmentation has been well explored and they can also provide essential inspirations for solving 3D tasks. Thus, for more adequate validation, we compare our method with four representative 2D methods by using their released source code, namely SPNet\cite{Xian_2019_CVPR}, ZS5Net\cite{NEURIPS2019_0266e33d}, PMOSR\cite{Zhang2021PrototypicalMA} and JoEm\cite{Baek_2021_ICCV}, on SemanticKITTI dataset. Some modifications are imposed on the source code for adapting to 3D point cloud data. We replace all 2D segmentation backbones with Cylinder3D\cite{zhu2020cylindrical}, the same as ours. Stacked calibration is applied in JoEm with $\gamma=0.08$ on softmax scores but is not used in SPNet due to terrible performance. The center loss is adopted instead of the BAR loss in JoEm since the feature interpolation is hard to implement in 3D sparse convolution. The results are shown in Table.~\ref{table:2D_compare}. Notably, these 2D methods have limited performance when adapting to 3D tasks with more complex 3D features.
We also try some generative methods~\cite{Cheng_2021_ICCV, Gu_2020} but fail for the similar reason that they usually rely on high-quality 2D feature map to train their generators, but 3D feature is challenging for generation. Our method is superior due to the rational exploration and utilization of each modal feature.

\begin{table}[t]
\caption{Comparision with other fusion methods on SemanticKITTI dataset with the 4-unseen-class setting.}
\resizebox{\linewidth}{!}{
\begin{tabular}{@{}lcccc}
\toprule
\multirow{2}{*}{Model} & \multirow{2}{*}{\begin{tabular}[c]{@{}c@{}}Seen\\ mIoU\end{tabular}} & \multirow{2}{*}{\begin{tabular}[c]{@{}c@{}}Unseen\\ mIoU\end{tabular}} & \multicolumn{2}{c}{Overall}   \\ \cline{4-5}  & & & \multicolumn{1}{c}{mIoU} & \multicolumn{1}{c}{hIoU} \\ \midrule
PointPainting\cite{Vora_2020_CVPR} & \textbf{58.9} & 16.3 & 49.9  & 25.5  \\
PointAugmenting\cite{wang2021pointaugmenting}&57.9 & 15.0 & 48.9  & 23.8 \\
2DPASS\cite{XuYan20222DPASS2P}  & 57.4 & 13.0 & 48.1  & 21.2 \\
PMF\cite{Zhuang_2021_ICCV}  & 56.9 & 14.1 & 47.9  & 22.6 \\
Deepfusion\cite{li2022deepfusion} & 54.9 & 15.8 & 46.7 & 24.5 \\
TransFuser \cite{9863660} & 54.1 & 11.9 & 45.2 & 19.5 \\
\midrule
Ours & 58.8 & \textbf{26.8} & \textbf{52.1} &\textbf{36.8} \\ \bottomrule
\end{tabular}
}
\label{table:compare_fusion}
\end{table}

\textbf{Comparison with popular multi-modal fusion methods.}
While it is intuitive that multi-sensor-based methods naturally outperform single-sensor-based methods since extra visual information is exploited, designing effective sensor-fusion methods for zero-shot tasks is non-trivial because we have to consider the complex projection relationship between semantic information and visual features. To verify the superiority of our multi-modal fusion approach, we apply two data-level fusion methods\cite{wang2021pointaugmenting, Vora_2020_CVPR} and four feature-level fusion methods\cite{XuYan20222DPASS2P,Zhuang_2021_ICCV, li2022deepfusion,9863660}, including two transformer-based methods\cite{li2022deepfusion, 9863660}, to the 3D zero-shot segmentation task by using the same baseline as our proposed method, which is a TGP\cite{chen2022zeroshot} models trained by ourselves for fair comparisons. As the results in Table. \ref{table:compare_fusion} shown, all previous camera-LiDAR-fusion methods gain limited or no improvements compared with the baseline because they fuse visual features directly without considering the semantic guidance, which is not suitable for zero-shot learning. In contrast, our method outperforms others by around $10\%$ since it allows semantic features to adaptively select valid LiDAR and image features for fusion, avoiding unnecessary information, and benefiting the knowledge transfer from seen classes to unseen classes. 

\begin{table}[t]
\begin{center}
\caption{Ablation experiments of the module of our framework on SemanticKITTI dataset with the 4-unseen-class setting.}
\label{table:ablation}
\resizebox{\linewidth}{!}{
\begin{tabular}{lcccc}
\toprule
\multirow{2}{*}{Model}       & \multirow{2}{*}{\begin{tabular}[c]{@{}c@{}}Seen\\ mIoU\end{tabular}} & \multirow{2}{*}{\begin{tabular}[c]{@{}c@{}}Unseen\\ mIoU\end{tabular}} & \multicolumn{2}{c}{Overall}   \\ \cline{4-5}  & & & \multicolumn{1}{c}{mIoU} & \multicolumn{1}{c}{hIoU} \\ \midrule
Ours                    & 58.8 & \textbf{26.8} 
                        & \textbf{52.1} & \textbf{36.8}                 \\ \hline
Ours w/o SGVF           & 58.8 & 23.4 & 51.3 & 33.5                     \\ \hline
Ours w/o SVFE           & \textbf{59.0} & 19.9 & 50.8 & 29.8                     \\ \hline
Ours w/o Image          & 58.3 & 20.0 & 50.2 & 29.8                     \\ \bottomrule
\end{tabular}
}
\end{center}
\end{table}

\subsection{Ablation Studies}
In this section, we conduct ablation studies on the SemanticKITTI dataset to verify the effectiveness of proposed modules in our network. Additionally, further analysis of the internal design of the SVFE and SGVF modules can be found in Appendix \ref{sec:abl_SVFE} and \ref{sec:abl_SGVF}. 


\textbf{Effect of SGVF.}
To verify the effectiveness of our feature-fusion strategy, we keep the backbone network and SVFE module and adopt a simple concatenation fusion instead of SGVF. We concatenate $F_{el}$ and $F_{ei}$ and utilize an MLP to compress the features to 128 dimensions. Then we perform visual-semantic alignment for the fused visual feature and $F_{es}$. As shown in Table.~\ref{table:ablation}, compared with using SGVF, the unseen mIoU drops about $3.5\%$, illustrating that SGVF fuses valid information and effectively transfers knowledge from seen classes to unseen ones.

\textbf{Effect of SVFE.}
To demonstrate the advantage of SVFE, we maintain the backbone network and the features extracted from the backbone are directly fed into SGVF. As Table.~\ref{table:ablation} shows, without SVFE, the unseen mIoU drops about $7\%$, showing that the huge semantic-visual gap leads to the difficulty of feature alignment in the joint space, while SVFE reduces the gap by feature enhancement.

\textbf{Effect of image modality.}
We also conduct ablation study for the image modality in Table.~\ref{table:ablation} by keeping the LiDAR backbone and its branch in the SVFE module. For semantic-visual alignment, we only utilize the single modal point cloud feature. We can see that compared with the multimodal setting, the unseen mIoU drops about $7\%$ without the appearance feature. Our method takes advantage of both sensors to match semantic feature space, which achieves significant improvement. It is worth noting that even with the LiDRA-only setting, our method is still superior to TGP (Table.~\ref{table:results}) due to our effective semantic-visual feature enhancement.

\section{Conclusions}
\label{sec:conclusions}
We make the first attempt to investigate the potential of multi-modal visual data in solving the transductive generalized zero-shot point cloud semantic segmentation problem. We have designed an effective multi-modal fusion method with mutual feature enhancement, which can adaptively determine what information should be taken from each modality under the semantic guidance for better semantic-visual alignment. Our method achieves SOTA performance on two large-scale datasets under diverse zero-shot settings.

\section{Acknowledgements}
This work was supported by NSFC (No.62206173), Natural Science Foundation of Shanghai (No.22dz1201900), MoE Key Laboratory of Intelligent Perception and Human-Machine Collaboration (ShanghaiTech University), Shanghai Frontiers Science Center of Human-centered Artificial Intelligence (ShangHAI), Shanghai Engineering Research Center of Intelligent Vision and Imaging.

{\small
\bibliographystyle{ieee_fullname}
\bibliography{paper}
}

\appendix
\label{sec:appendix}
\clearpage
\begin{appendices}
\section{More Details of SVFE}
\label{sec:abl_SVFE}

\textbf{Why SVFE improves the performance?} 
The main function of the SVFE module is to narrow the semantic-visual gap and facilitate early knowledge transfer between semantic and visual spaces, rather than simply scaling up the model. To demonstrate the importance of the semantic-visual interaction, we conduct an experiment where we replace it with self-attention operation with the same parameter scale for each single modality. The results in Table.\ref{tab: svfe} show the performance drops sharply without the SVFE module.

\begin{table}[h]
\caption{Ablation experiments of the design of SVFE module on SemanticKITTI dataset with the 4-unseen-class setting,}
\label{tab: svfe}
\resizebox{\linewidth}{!}{
\begin{tabular}{lcccc}
\hline
\multirow{2}{*}{Model}       & \multirow{2}{*}{\begin{tabular}[c]{@{}c@{}}Seen\\ mIoU\end{tabular}} & \multirow{2}{*}{\begin{tabular}[c]{@{}c@{}}Unseen\\ mIoU\end{tabular}} & \multicolumn{2}{c}{Overall}   \\ \cline{4-5}  & & & \multicolumn{1}{c}{mIoU} & \multicolumn{1}{c}{hIoU} \\ \hline
baseline                           & 54.6 & 17.3 & 46.7 & 26.3                  \\ 
baseline + self attention           & 57.3 & 19.4 & 49.3 & 29.0                  \\
baseline + SVFE           & \textbf{58.8 }& \textbf{23.4} & \textbf{51.3} & \textbf{33.5}                     \\
\hline
image features first($F_{es}'$)            & 58.3 & 16.1 & 49.4  &  25.2 \\
point cloud features first($F_{es}$)     & \textbf{58.8} & \textbf{26.8} & \textbf{52.1} & \textbf{36.8} \\  \bottomrule
\end{tabular}
}
\end{table}

 \textbf{Does fusion order in SVFE matter?}
As mentioned in Sec 3.4, semantic feature enhancement is implemented as: $F_{es} = \mathrm{TD}(\mathrm{TD}(F_{s}, F_{l},F_{l}),F_{i},F_{i})$.
We provide the result of fusing image visual features first and then point cloud visual features:
$F_{es}' = \mathrm{TD}(\mathrm{TD}(F_{s}, F_{i},F_{i}),F_{l},F_{l})$. As shown in Table.\ref{tab: svfe}, The ordering of feature fusion presented in the paper is superior because visual features extracted from point clouds are more central to 3D point cloud segmentation. By fusing these visual features with semantic features first, we are able to provide better guidance for the segmentation process.

\section{More Details of SGVF}
\label{sec:abl_SGVF}
 \textbf{Are there any better fusion methods than SGVF module?}
As SGVF adopts an attention-based design, to further validate the effectiveness of the SGVF module, we design experiments to compare our method with two variants of transformer-based multimodal fusion methods, as shown in Table.\ref{tab:sgvf}. We find that the performance of ``w/o SGVF, w/ cross attention``, which uses LiDAR to query image features for fusion without considering semantic features, is not as good as our SGVF module. This is consistent with our intuition that simply fusing the visual features without considering the semantic information is not sufficient for zero-shot tasks. However, the result of ``w/ SGVF, w/ self attention`` is unexpected. The performance of the method with the added self-attention mechanism for the fused features is lower than that of SGVF, even though the parameter quantity is increased. This suggests that simply increasing the model complexity does not necessarily lead to better performance. In fact, the additional self-attention mechanism may have introduced noise and decreased the discriminative power of the fused features.

\begin{table}[h]
\caption{Ablation experiments of the design of SGVF module on SemanticKITTI dataset with the 4-unseen-class setting,}
\label{tab:sgvf}
\resizebox{\linewidth}{!}{
\begin{tabular}{@{}lcccc}
\hline
\multirow{2}{*}{Model}       & \multirow{2}{*}{\begin{tabular}[c]{@{}c@{}}Seen\\ mIoU\end{tabular}} & \multirow{2}{*}{\begin{tabular}[c]{@{}c@{}}Unseen\\ mIoU\end{tabular}} & \multicolumn{2}{c}{Overall}   \\ \cline{4-5}  & & & \multicolumn{1}{c}{mIoU} & \multicolumn{1}{c}{hIoU} \\ \hline
w/o SGVF, w/ cross attention   & 56.6 & 21.9 & 49.3 & 31.6 \\
w/ SGVF, w/ self attention   & 50.4 & 21.2 & 44.3 & 29.8 \\\hline
Ours & \textbf{58.8} & \textbf{26.8} & \textbf{52.1} &\textbf{36.8} \\
\bottomrule
\end{tabular}
}
\end{table}

\section{Model inference time} 
\label{sec:abl_infer_time}
 With the addition of an extra image modality, our model's inference time is \textbf{0.097 seconds per frame}, slightly larger than \textbf{0.087s/f} of the SOTA single-modal method TGP[13]. But our model outperforms it with more than \textbf{$50\%$} improvement of unseen category mIOU. Furthermore, it yields real-time performance (All of the results are tested on 1 NVIDIA GTX3090 GPU).

\section{The impact of various image encoders on performance}
\label{sec:abl_image_encoder}
 We employed ResUnet-34 as our image encoder (L591). To show the impact of various image encoders, we replace the encoder with ResUnet-18 and ResUnet-50 and get comparable performance, as shown in the below table.

\begin{table}[h]
\begin{center}
\begin{tabular}{lcccc}
\toprule
\multirow{2}{*}{Model}       & \multirow{2}{*}{\begin{tabular}[c]{@{}c@{}}Seen\\ mIoU\end{tabular}} & \multirow{2}{*}{\begin{tabular}[c]{@{}c@{}}Unseen\\ mIoU\end{tabular}} & \multicolumn{2}{c}{Overall}   \\ \cline{4-5}  & & & \multicolumn{1}{c}{mIoU} & \multicolumn{1}{c}{hIoU} \\ \midrule
ResUnet-18   & 57.3 & 24.7 & 50.4 & 34.5         \\
ResUnet-50   & \textbf{58.9} & \textbf{27.1} & \textbf{52.2} & \textbf{37.1} \\ \midrule
Ours(ResUnet-34)     & 58.8 & 26.8 & 52.1 & 36.8  \\ \bottomrule
\end{tabular}
\end{center}
\end{table}

\section{Discussion on the CLIP Model}
\label{sec:abl_clip}
\begin{table}[t]
\begin{center}
\caption{CLIP model experiment on SemanticKITTI dataset with the 4-unseen-class setting.}
\label{table:clip_model}
\resizebox{\linewidth}{!}{
\begin{tabular}{lcccc}
\toprule
\multirow{2}{*}{Model}       & \multirow{2}{*}{\begin{tabular}[c]{@{}c@{}}Seen\\ mIoU\end{tabular}} & \multirow{2}{*}{\begin{tabular}[c]{@{}c@{}}Unseen\\ mIoU\end{tabular}} & \multicolumn{2}{c}{Overall}   \\ \cline{4-5}  & & & \multicolumn{1}{c}{mIoU} & \multicolumn{1}{c}{hIoU} \\ \midrule
Ours $\leftarrow$ CLIP model   & 56.6 & 14.1 & 47.7 & 22.6          \\
Ours     & \textbf{58.8} & \textbf{26.8} & \textbf{52.1} & \textbf{36.8}  \\ \bottomrule
\end{tabular}
}
\end{center}
\end{table}

Given the success of the CLIP \cite{CLIP} model in 2D zero-shot segmentation \cite{Xu2021clip1, Lddecke2022clip2, Ding2022Decouplingclip3, Li2022LanguagedrivenSS, zhou2022extract}, we aim to investigate its potential for 3D point cloud semantic segmentation by incorporating the CLIP model into our method. We follow the approach used in MaskCLIP \cite{zhou2022extract}, where the class name is inserted into 85 hand-crafted prompts and they are fed into CLIP's text encoder to generate multiple text features. Additionally, we replace the 2D ResUNet backbone with MaskCLIP+.
As shown in Table \ref{table:clip_model}, even though unseen objects may already occur in the CLIP training data, causing data leakage, the incorporation of the CLIP model still performs worse than our \textbf{pure zero-shot method}.
It is mainly because CLIP is based on the contrastive learning between image and text pairs and the significant disparity between point cloud features and image features makes point cloud visual features difficult to align with semantic features extracted by CLIP. 
However, it is interesting to explore the projection between point cloud and images and transfer the knowledge learnt by CLIP to solve 3D zero-shot problems in large scenarios.

\end{appendices}

\end{document}